\documentclass[]{OAGM}
\OAGMarXiv{1404.3538}

\usepackage{float} % fix the table
\usepackage{bold-extra}
\usepackage{amsmath,bm}
\usepackage{color}
\usepackage{amsmath,amsopn}
\usepackage{bbm}
\usepackage{mathrsfs}
\usepackage{tikz,pgfplots}
\usepackage{listings}
\usepackage{color}
\usepackage{caption}
\usepackage{subcaption}
\usepackage{epstopdf}
\usepackage{mathtools} 
\usepackage{tabularx}
\usepackage{booktabs}
\usepackage[square, comma, sort&compress, numbers]{natbib}
\usepackage{graphics}
\usepackage{amsfonts}
\usepackage{mathtools}

\mathtoolsset{showonlyrefs}
\let\originaleqref=\eqref
\renewcommand{\eqref}{Equation~\originaleqref}

\DeclareMathOperator*{\newarg}{arg} 
\newcommand{\avg}[1]{\left\langle #1 \right\rangle}           % for average
\newcommand{\norm}[1]{\left\lVert #1 \right\rVert}            % for norm
\newcommand{\bra}[1]{\left\{ #1 \right\}}            % for {}
\newcommand{\sbra}[1]{\left[ #1 \right]}            % for []
\newcommand{\lbra}[1]{\left( #1 \right)}            % for ()

\title{A Comparison of First-order Algorithms for Machine Learning}

\author{Wei Yu Thomas Pock\\
  Computer Graphics and Vision, Graz University of Technology, Austria}
  
\begin{document}
\maketitle

\begin{abstract}
Using an optimization algorithm to solve a machine learning problem is one of mainstreams in the field of science. In this work, we demonstrate a comprehensive comparison of some state-of-the-art first-order optimization algorithms for convex optimization problems in machine learning. We concentrate on several smooth and non-smooth machine learning problems with a loss function plus a regularizer. The overall experimental results show the superiority of primal-dual algorithms in solving a machine learning problem from the perspectives of the ease to construct, running time and accuracy.  \end{abstract}

\section{Introduction}

Optimization is the key of machine learning. Most machine learning problems can be cast as optimization problems. Furthermore, practical applications of machine learning usually involve a massive and complex data set. Thus, efficiency, accuracy and generalization of the optimization algorithm (solver) should be regarded as a crucial issue \citep{Bennett:2006:IOM:1248547.1248593}. Many papers present dedicated optimization algorithms for specific machine  learning problems. However, little attention has been devoted to the ability of a solver for a specific class of machine learning problems. The most common structure of machine learning problems is a loss function plus a regularizer. The loss function calculates the disparity between the prediction of a solution and the ground truth. This term usually involves the training data set. For example, the well known square loss is for the purpose of regression problems and hinge loss is for the purpose of maximum margin classification. The regularizer usually uses a norm function. For example, group lasso is an extension of the lasso for feature selection. It can lead to a sparse solution within a group. 

In general, we consider convex optimization problems of the following form
$$\begin{cases} \mathrm{minimize} & E(x)=F(x)+\lambda G(x)\\ \mathrm{such\ that}& x\in C\end{cases}$$ where $F$ and $G$ are continuous, convex functions and $C$ is a convex set. $E$ denotes the energy of a machine learning problem. By convention, $F$ usually denotes a loss function and $G$ denotes a regularization term. $\lambda$ is a parameter controlling the tradeoff between a good generalization performance and over-fitting. This kind of problems frequently arise in machine learning. A substantial amount of literature assumes that either $F$ or $G$ is smooth and cannot be used to optimize the case where $F$ and $G$ are both non-smooth. 

Some solvers provide an upper bound N on the number of iterations n such that $E^n-\widehat{E}\le e, n\ge N$, where $e$ is an error tolerance and $\widehat{E}$ is the minimum of $E$. Sometimes this estimation is too pessimistic which means the resultant $N$ is excessive large. In this case, it is hard to evaluate the performance of a solver by this upper bound. On the other side, convergence rate describes the speed of converging when a solver approaches the optimal solution. But it is unpredictable to know when $n$ is large enough. Therefore, the performance of solvers is still difficult to tractable. 

In this paper, we compare four state-of-the-art first-order solvers (Fobos\cite{Duchi:2009:EOB:1577069.1755882}, FISTA\cite{Beck:2009:FIS:1658360.1658364}, OSGA\cite{Osga} and primal-dual algorithms\cite{raey}) by the following properties: convergence rate, running time, theoretically known parameters, robustness in practice for machine learning problems. We present tasks within dimensionality reduction via compressive sensing, SVMs, group lasso regularizer for grouped feature selection, $\ell_{1,\infty}$ regularization for multi-task learning, trace norm regularization for max-margin matrix factorization. The last three machine learning problems are chosen from \cite{Yangarxivpd}. Unlike other literature which plots energy versus the number of iterations, in this paper we illustrate the results by log-log figures which clearly show the convergence rate in applications of machine learning. 

The paper is organized as follows. Section ~\ref{sec:The solvers} introduces four solvers. Then it summarizes primal-dual algorithm of Chambolle and Pock \cite{raey} and describes heuristic observations. Section ~\ref{sec:The machine learning problem} gives an introduction about the general structure (a loss function plus a regularizer) of machine learning problems we focus on in this paper. Section ~\ref{sec:Result} demonstrates the performance of different solvers and the conclusion is presented at the end.

\section{Solvers}
\label{sec:The solvers}
\subsection{Review}

Fobos \cite{Duchi:2009:EOB:1577069.1755882} and fast iterative shrinkage-thresholding algorithm \cite{Beck:2009:FIS:1658360.1658364} (FISTA) aim to solve a convex problem which is a sum of two convex functions. Neumaier \cite{Osga} proposes a fast subgradient algorithm
with optimal complexity for minimizing a convex function named optimal subgradient algorithm (OSGA). Chambolle and Pock \cite{raey} propose a primal-dual algorithm (hereinafter referred as PD CP) and applied it to several imaging problems. Tianbao Yang et al. \cite{Yangarxivpd} propose another primal-dual algorithm and applied it to machine learning tasks. However, PD CP is more general in the following aspects. The step size of PD CP is $\sqrt{2}$ times larger than \cite{Yangarxivpd} and PD CP makes steps in both primal and dual variables. 

\begin{table}[bhp]
%\vspace{0.3cm}
\begin{tabularx}{\textwidth}{ccccc}
\toprule
Solver    &  Convergence rate   & $F$   &   $G$ & $E$  \\\midrule
Fobos  & $O(1/\sqrt{n})$ & convex& convex &-\\\midrule
FISTA &$O(1/n^2)$ & $C^{1,1}$ &convex &-\\\midrule
OSGA & $O(1/\sqrt{n})$ &-&-&convex  \\\midrule  % \\\bottomrule
PD CP & $O\lbra{1/n}$ &convex&convex&- \\\bottomrule 
\end{tabularx}
%\vspace{-0.3cm}
\caption{Tab Comparison of solvers}\label{Tab Comparison of solvers}
%\vspace{0.3cm}
\end{table}

We summarize four solvers by Table ~\ref{Tab Comparison of solvers}. Each solver can achieve the convergence rate under the property of $F, G, E$ given by each row of Table ~\ref{Tab Comparison of solvers}. When we solve machine learning problems using four solvers, we need to set the value of parameters and format the machine learning problems to a suitable model for a solver. Setting the initial step size C in Fobos for a problem with non-smooth function is an open question. For PD CP, we do not know the best ratio $a=\sqrt{\tau/\sigma}$. When solving a problem with a non-smooth function by FISTA, we have to smooth the non-smooth loss function. This rises a problem of selecting the value of smoothness parameter $\epsilon$ in smoothing techniques. To make the comparison convincing, we examine several values for the above three parameters $C, a, \epsilon$ and choose the best one.

\subsection{The general PD CP}

In this section, we review the primal dual algorithm proposed in \citep{raey}. Let $X,Y$ be two finite dimensional real vector spaces with an inner product $\avg{\cdot,\cdot}$ and norm $\norm{\cdot}=\avg{\cdot,\cdot}^{\frac{1}{2}}$. The map $K:X\rightarrow Y$ is a continuous
linear operator with induced norm\begin{equation*}\norm{K}=
\max\bra{\norm{Kx}_2:x\in X,\ \norm{x}_2\le 1}. \end{equation*} PD CP is to solve the generic saddle-point problem
\begin{equation}\min\limits_{x\in X}\max\limits_{y\in Y}\bra{\avg{Kx,y}+G(x)-F^*(y)}\label{eq saddle}\end{equation} where $x$ is the primal variable and $y$ is the dual variable. $G$ and $F^*$ both are proper convex, lower-semicontinuous functions. The primal form of ~\eqref{eq saddle} is $\min\limits_{x\in X}F(Kx)+G(x)$. To introduce the dual variable y, one way is using Lagrange multipliers, e.g., applicable when the function represents hard constraints. The other way is to calculate the convex conjugate $F^*$ of loss function $F$. Then the loss function can be expressed by its convex conjugate. 
Take the hinge loss for example. We simplify the definition of hinge loss as $f(z)=\norm{\max(0,z)}_1,z \in \mathbb{R}^N$. Its convex conjugate is,
\begin{equation*}
f^*(y)=\begin{cases} 0& y\in P\\
+\infty & y\notin P
\end{cases}
\end{equation*}
where $P=\bra{y\in \mathbb{R}^N:\forall y_i\in \sbra{0,1}}$, $y_i$ is the $i^{th}$ component of $y$.
Let $z=1-Kx, x\in \mathbb{R}^d$ and $K \in \mathbb{R}^{N\times d}$. Then according to $f(z)=\max\limits_y\avg{z,y}-f^*(y)$, $f(1-Kx)$ can be defined as,
$$f(1-Kx)=\max\limits_y\avg{1-Kx,y}-f^*(y).$$
Therefore in the primal-dual model, $F^*$ should be formulated as $-\sum\limits_{i=1}^Ny_i+f^*(y)$.

Before summarizing PD CP, we introduce the proximal operator. Let $G:\ X\rightarrow \mathbb{R}\cup\bra{+\infty}$ be a proper convex, lower-semicontinuous function. The proximal operator of $G$ with parameter $\tau$ is defined by
\begin{align}
\mathrm{prox}_{\tau G}(v)&=(I+\tau\partial G)^{-1}(v)\\
&=\newarg\min\limits_{x\in X}(\tau G(x)+\frac{\norm{x-v}_2^2}{2})
\end{align}
Since Euclidean norm is strong convex, $\mathrm{prox}_{\tau G}(v)$ is unique. PD CP proceeds by iteratively maximizing with respect to the dual variable and minimizing with respect to the primal variable by proximal operators.

Now we summarize PD CP as follows,
\vspace{0.1cm}
\\
\fbox{
\begin{minipage}[c]{\textwidth}%
$\bullet$ Initialization: $\tau\sigma\le\frac{1}{\norm{K}^2}$,$\theta\in\sbra{0,1}$,$(x^0,y^0)\in X\times Y$,$\overline{x}^0 = x^0, \lambda \in \mathbb{R}$

$\bullet$ Iterations $n\ge 0$: Update $x^n, y^n$ as follows,
\begin{align}
   y^{n+1}&=(I+ \sigma \partial F^*)^{-1}(y^n+\sigma K \overline{x}^n) \label{eq pd1} \\
   x^{n+1}&=(I+\tau \lambda\partial G)^{-1}({x}^n-\tau K^*y^{n+1})  \label{eq pd2} \\
      \overline{x}^{n+1} &= x^{n+1}+\theta (x^{n+1}-x^{n})
\end{align}
 \end{minipage}
}
\vspace{0.1cm}

PD CP conducts proximal operators of $\sigma F^*$ and $\tau \lambda G$ respectively and then PD CP makes its scheme semi-implicit by letting $\overline{x}^{n+1} = x^{n+1}+\theta (x^{n+1}-x^{n})$. This operation equals to making one more step in the direction of $x^{n+1}-x^{n}$. We refer to \citep{raey} for more information. The condition of the convergence of PD CP is $\tau\sigma\le\tfrac{1}{\norm{K}^2}$.

\subsection{Heuristics for the primal dual algorithm}

In this section, we introduce two heuristic observations of the primal-dual algorithm proposed by Chambolle and Pock \cite{raey}.  

\subsection{Heuristics for the ratio of the primal step size to the dual step size}

In PD CP, we define $a=\sqrt{\tau/\sigma}$. 
 How to choose $a$ to achieve the best performance is still an unsolved problem. However, Chambolle and Pock \cite{raey} show that
 \begin{equation}
 \begin{aligned}
 \sbra{\avg{Kx_N,\widehat{y}}-F^*(\widehat{y})+G(x_N)}-&\sbra{\avg{K\widehat{x},y_N}-F^*(y_N)+G(\widehat{x})}\\
 &\le\frac{1}{N}(\frac{\norm{\widehat{y}-y^0}^2}{2\sigma}+\frac{\norm{\widehat{x}-x^0}^2}{2\tau})
 \label{eq lower}\end{aligned}
 \end{equation}
where $x_N=(\sum\limits_{n=1}^Nx^n)/N$, $y_N=(\sum\limits_{n=1}^Ny^n)/N$ and $(\widehat{x}, \widehat{y})$ is the saddle point. The RHS of ~\eqref{eq lower} is non-negative because
\begin{align}
\sbra{\avg{Kx_N,\widehat{y}}-F^*(\widehat{y})+G(x_N)}&\ge\sbra{\avg{K\widehat{x},\widehat{y}}-F^*(\widehat{y})+G(\widehat{x})}\\
&\ge  \sbra{\avg{K\widehat{x},y_N}-F^*(y_N)+G(\widehat{x})}.
\end{align}  And when $(x_N,y_N)$ is a saddle point, the LHS of ~\eqref{eq lower} equals zero. To minimize the upper bound of $\sbra{\avg{Kx_N,\widehat{y}}-F^*(\widehat{y})+G(x_N)}-\sbra{\avg{K\widehat{x},y_N}-F^*(y_N)+G(\widehat{x})}$, we plug $\tau=\frac{1}{\norm{K}^2\sigma}$, the largest value that guarantees convergence, into the RHS of ~\eqref{eq lower} and get
\begin{equation}\frac{1}{N}(\frac{\norm{\widehat{y}-y^0}^2}{2\sigma}+\frac{\norm{\widehat{x}-x^0}^2\norm{K}^2\sigma}{2}).\label{eq: a}\end{equation}
~\eqref{eq: a} is a convex function of $\sigma$. We take the derivative of ~\eqref{eq: a} with respect to $\sigma$,
$$\sigma=\frac{\norm{\widehat{y}-y^0}}{\norm{\widehat{x}-x^0}\norm{K}}.$$
Thus we can conclude that $\frac{1}{N}(\frac{\norm{\widehat{y}-y^0}^2}{2\sigma}+\frac{\norm{\widehat{x}-x^0}^2}{2\tau})$ reaches its minimum when $a=\sqrt{\frac{\tau}{\sigma}}=\frac{\norm{\widehat{x}-x^0}}{\norm{\widehat{y}-y^0}}$. However, $\widehat{x}$ and $\widehat{y}$ are not available because they are what we want to calculate. 

\subsection{Heuristics for the adaption of step sizes}

We observe that the convergence condition \cite{raey} $\tau\sigma\le\tfrac{1}{\norm{K}^2}$ can be relaxed to accelerate the algorithm. We refer to the resulting scheme as Online PD CP. Although Online PD CP converges in the experiments of this paper, we do not prove its convergence theoretically. Online PD CP try to seek a larger step size. Once it finds one, it is faster than PD CP in the experiments of this paper. The difference between PD CP and Online PD CP is that Online PD CP starts with a larger step size ($\tau\sigma>\frac{1}{\norm{K}^2}$) and decreases it according to a certain rule. We employ the following scheme,

\vspace{0.1cm}
 \begin{equation}
  \left\{
   \begin{array}{rl}
\widetilde{L}^{n+1}&=\frac{\avg{K(x^n-x^{n-1}),y^{n+1}-y^n}}{\norm{x^{n-1}-x^n}\norm{y^{n+1}-y^n}}\\
   L^{n+1}&=\max\lbra{L^n,\widetilde{L}^{n+1}}\\
    \tau^{n+1}&=\frac{a}{{L^{n+1}}}, \sigma^{n+2}=\frac{1}{a{L^{n+1}}} \\
\end{array}
  \right.
  \label{eq L} \end{equation}
\vspace{0.1cm}

Thus how to choose a proper $L$ is the main concern of Online PD CP. As shown in ~\eqref{eq L}, we let $L^{n+1}=\max\lbra{L^n,\widetilde{L}^{n+1}}$. If $L^n<\widetilde{L}^{n+1}$, we increase $L^{n+1}$ to $\widetilde{L}^{n+1}$. Thus, $\norm{K}$ is a upper bound of $L$. Chambolle and Pock \cite{raey} proves the convergence when $L=\norm{K}$.

Another observation is that the larger step size may lead to a large $L$. If $\widetilde{L}^{n+1}$ is smaller than $L^{n}$, Online PD CP does not update $L^{n+1}$. It is a sign of convergence and stability. That is, inappropriate large step sizes lead to divergence and a large $\widetilde{L}$ which may be close to $\norm{K}$. This obeys the principle of Online PD CP. To explore more possible step sizes, we decrease the step size to an appropriate degree rather than choose the maximum between $L^n$ and $\widetilde{L}^{n+1}$. This is the reason that we smooth $L$. We can devise different rules to smooth $L$. For example, we let $L=(L+\kappa\max(L,{L}^{n+1} ))/(1+\kappa), \kappa>0$. We set $\kappa=0.618$ for all experiments. The other use of Online PD CP is the case that $\norm{K}$ is non-calculable, e.g., $K$ is not known explicitly.

\section{The machine learning problems}

\label{sec:The machine learning problem}
Machine learning problems in this paper can be formulated as a convex minimization problem consisting of a loss function $F$ and a regularizer $G$. We summarize machine learning problems in Table 2. In each row of table 2, the last two columns show the loss function and regularizer we used in the machine learning problem given by column 1. For more information about each machine learning problem, refer to the literature given by column 2.

In Table 2, $Q, H, \hat{H}$ are positive-semidefinite matrices and $\delta$ is an indicator function. In experiment of Kernel SVM, we calculate the dual form of the primal form \cite{Chapelle2007} such that $F$ becomes a smooth convex function as shown in row 5 of Table ~\ref{tab ML}. And we solve this dual form by Fobos, FISTA and PD CP. Because this dual form is a constrained optimization problem which is not easy to be solved by OSGA, we use OSGA to solve the primal form \cite{Chapelle2007} as shown in row 4 of Table ~\ref{tab ML}.

\vspace{0.3cm}
\begin{tabularx}{\textwidth}{cccc}
\toprule
Machine learning problem    &   Ref.              & $F$  &   $G$\\\midrule
Dimensionality Reduction\footnote{MNIST is available at http://yann.lecun.com/exdb/mnist/.\label{MNIST}}  &   \citep{Gao:2012:DRV:2206432.2206550}                          & square  & $\ell_{2,1}$\\\midrule
Linear SVM\footnote{`svmguide1' is available at http://www.csie.ntu.edu.tw/~cjlin/libsvmtools/datasets/.\label{svmguide1}}                 &  \citep{ICML2013_pele13} & hinge  & $x^TQx$  \\ \midrule
Kernel SVM\textsuperscript{\ref{svmguide1}}                & \citep{Chapelle2007}    & hinge  &$x^THx$\\\midrule
Kernel SVM \textsuperscript{\ref{svmguide1}}       & -     & $x^T\hat{H}x$ &$-\sum x_i+\delta (x)$\\\midrule
Feature Selection\footnote{MEMset Donar is available at http://genes.mit.edu/burgelab/maxent/ssdata/.}    & \citep{DBLP:conf/icml/YangXKL10} &absolute loss &  group lasso\\ \midrule
Multi-Task Learning \textsuperscript{\ref{MNIST}}   & \citep {Yangarxivpd} &$\epsilon$-insensitive &  $\ell_{1,\infty }$ \\\midrule
Matrix Factorization\footnote{`100K MovieLens' is available at http://www.grouplens.org/node/12.}     & \citep {Yangarxivpd}&hinge & trace norm\\\bottomrule
\end{tabularx}
\vspace{-0.3cm}
\captionof{table}{Machine learning problems}\label{tab ML}

\section{Results}
\label{sec:Result}
\subsection{Experimental settings}
When $\nabla F$ is with a Lipschitz constant $L$, the convergence rate of Fobos can be $O(1/n)$ \citep{Beck:2009:FIS:1658360.1658364}. Thus, we only compare Fobos with FISTA, OSGA and PD CP in the machine learning problems where $F$ and $G$ are both non-smooth. In all experiments, we initialize the primal variable and the dual variable to a null vector. All algorithms were implemented in Matlab and executed on a 2.66 GHz CPU,
running a 64 Bit Windows system. 

\subsection{Covergence and time comparison}
\begin{figure}[ht!] % "[t!]" placement specifier just for this example
\begin{subfigure}{0.5\textwidth}
\includegraphics[width=\linewidth]{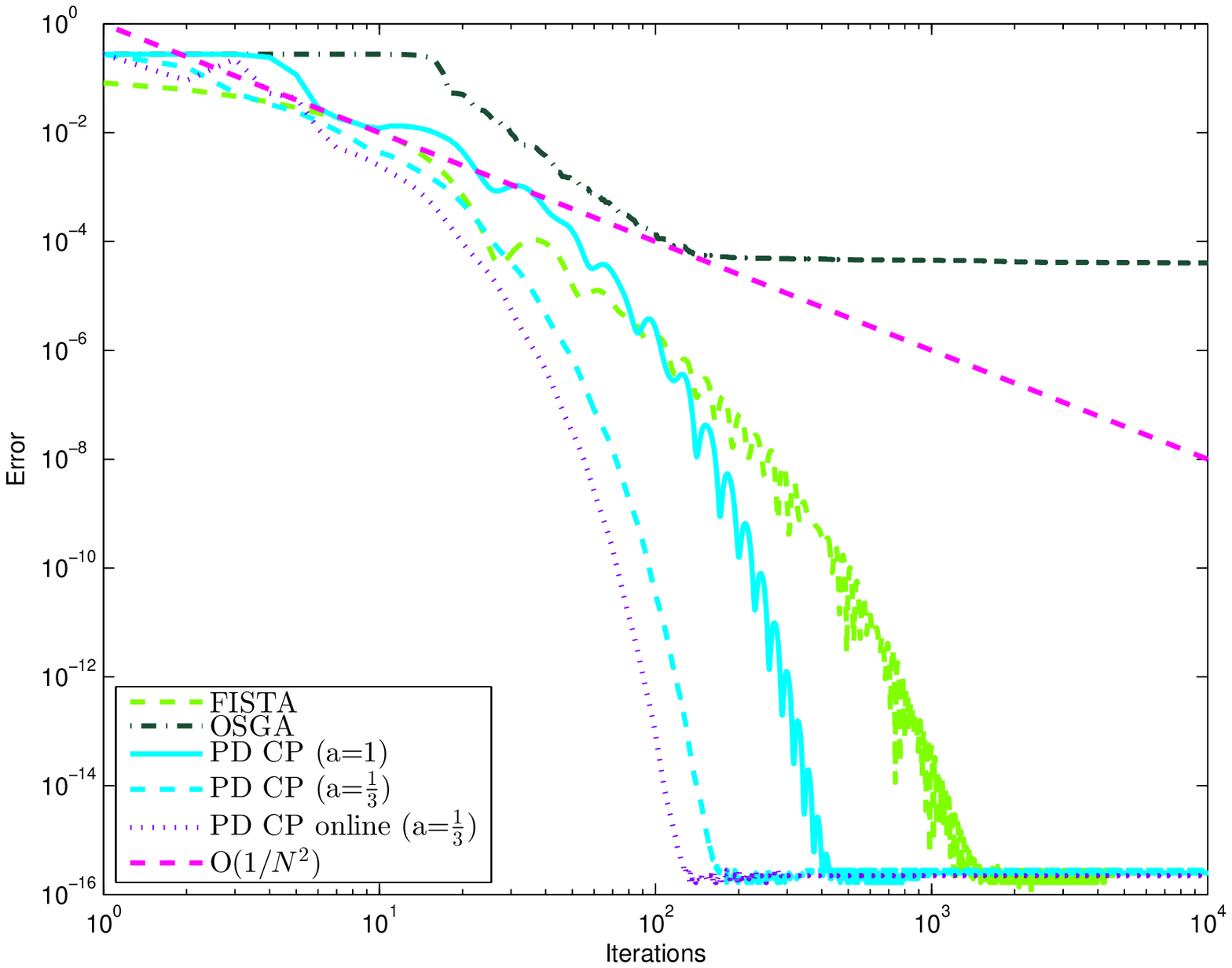} 
  \caption{Dimensionality reduction using $\lambda=1$}
     \label{fig cs}
\end{subfigure}\hspace*{\fill}
\begin{subfigure}{0.5\textwidth}
\includegraphics[width=\linewidth]{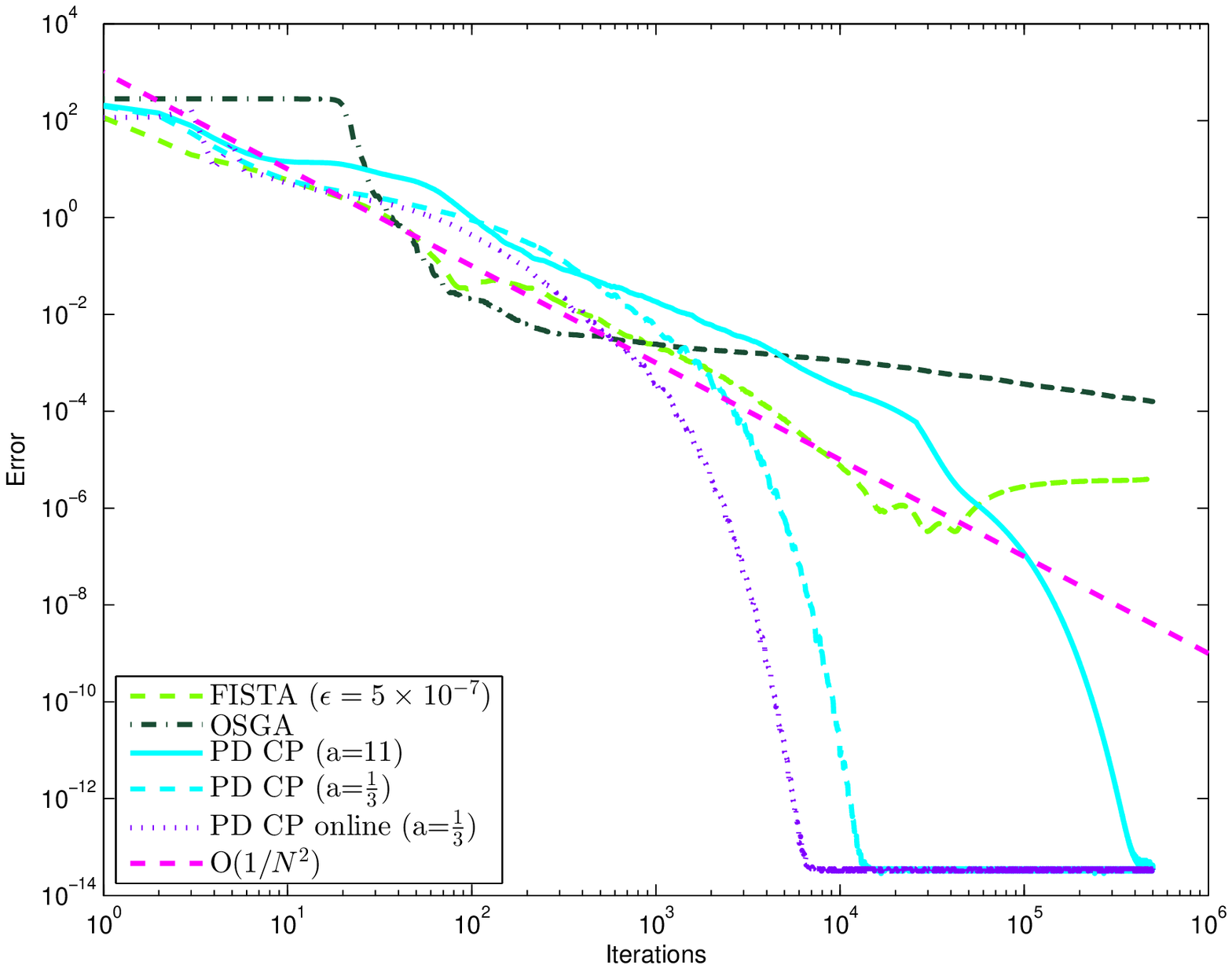}
 \caption{Linear SVM using $\lambda=10$} \label{fig linear svm}
\end{subfigure}

\medskip
\begin{subfigure}{0.5\textwidth}
\includegraphics[width=\linewidth]{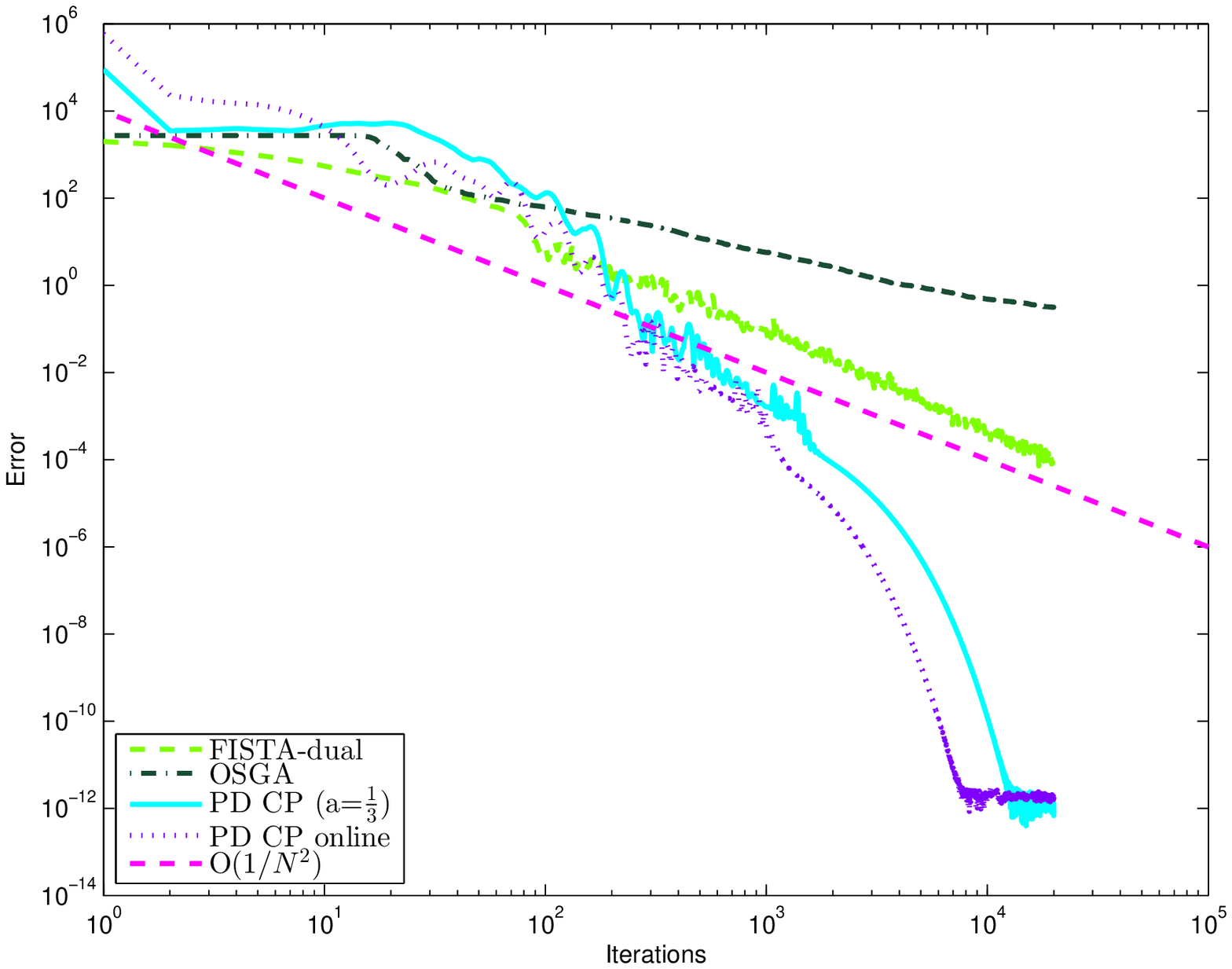}
\caption{Kernel SVM using $\lambda=1$} \label{fig kernel svm}
\end{subfigure}\hspace*{\fill}
\begin{subfigure}{0.5\textwidth}
\includegraphics[width=\linewidth]{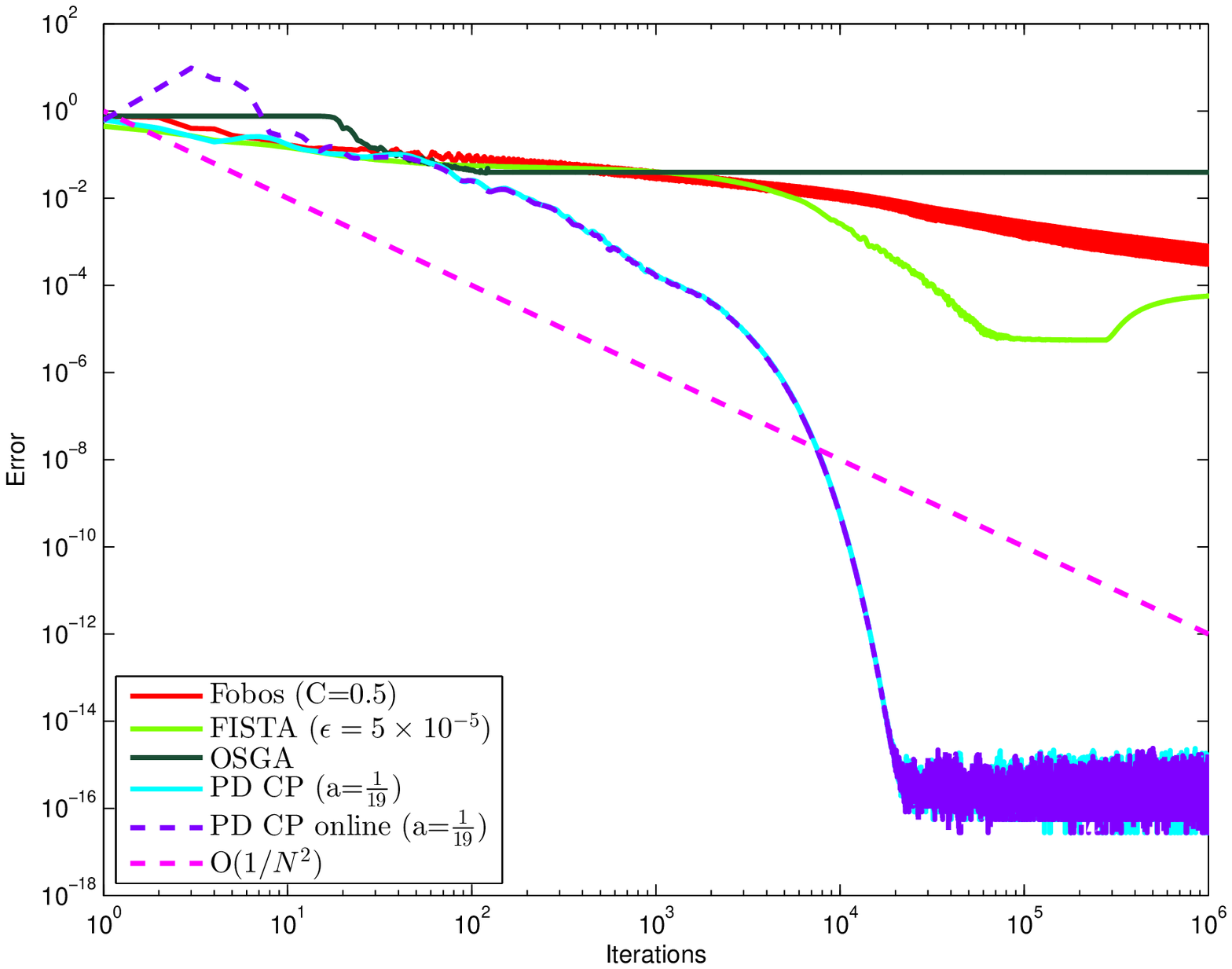}
 \caption{Feature Selection using $\lambda=10^{-3}$}
        \label{fig:fig GL a}
\end{subfigure}

\medskip
\begin{subfigure}{0.5\textwidth}
\includegraphics[width=\linewidth]{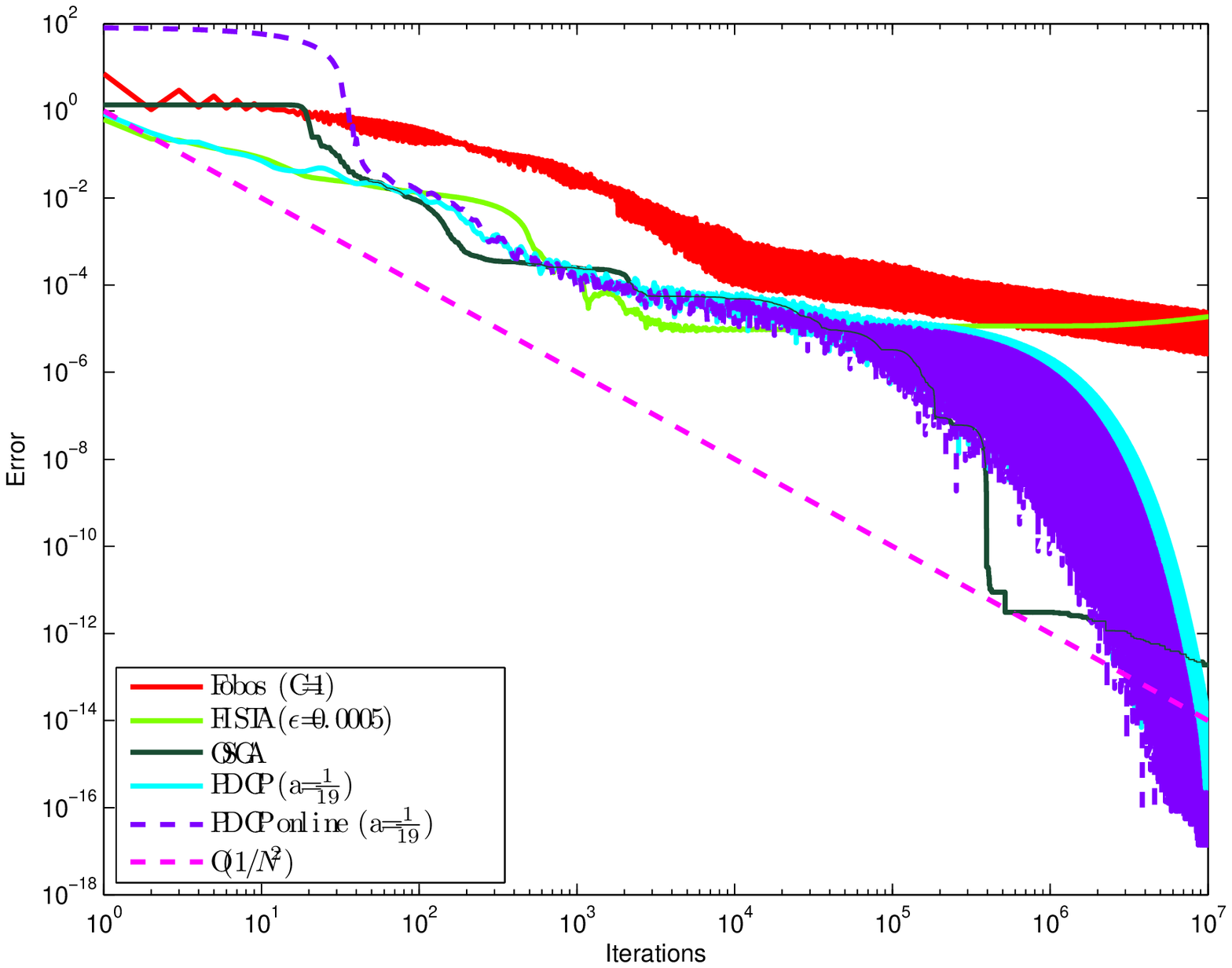}
  \caption{Multi-Task Learning using $\lambda=10^{-3}$}
        \label{fig MTL c}
\end{subfigure}\hspace*{\fill}
\begin{subfigure}{0.5\textwidth}
\includegraphics[width=\linewidth]{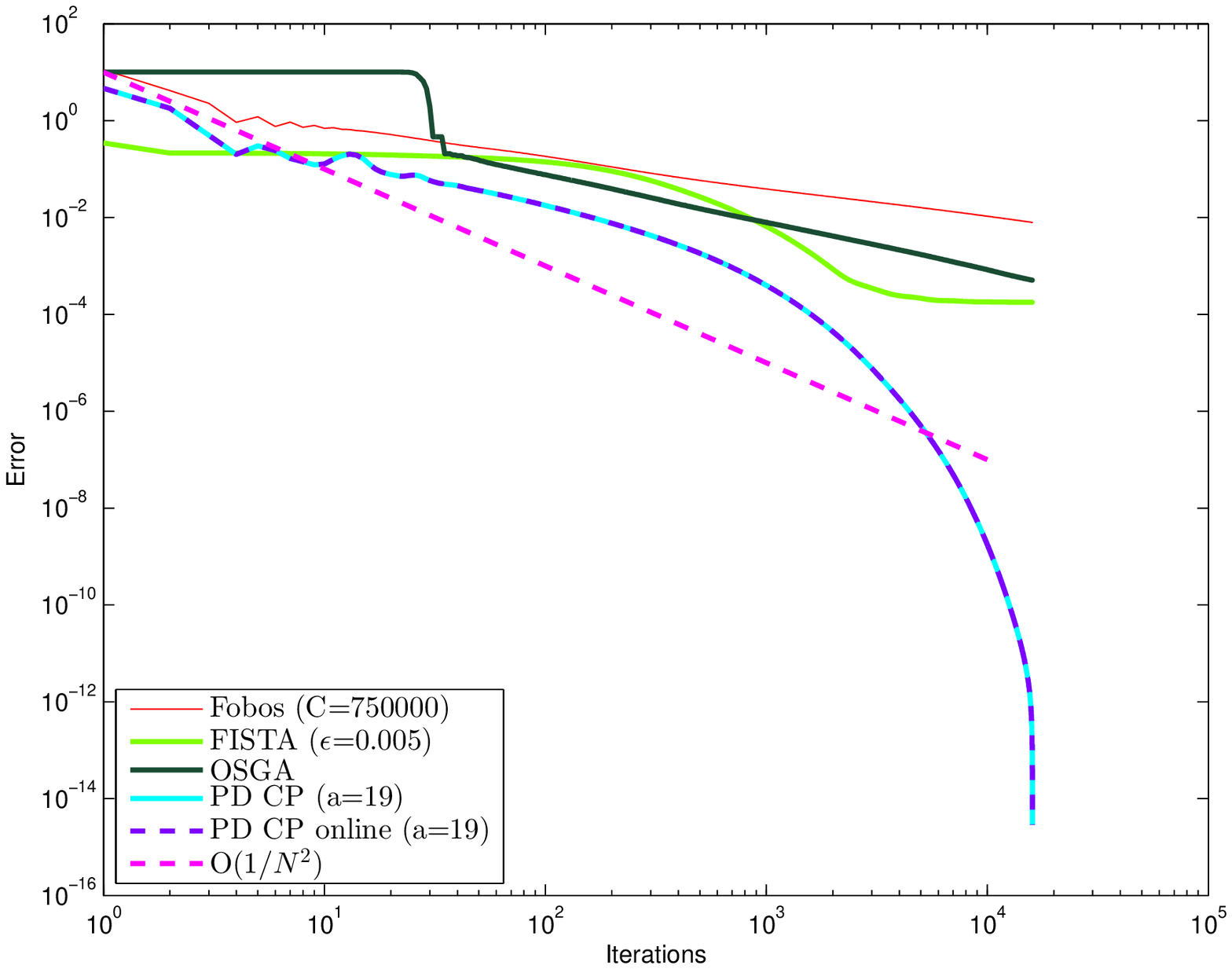}
\caption{Matrix Factorization using $\lambda=10^{-5}$}
        \label{fig mmmf d}
\end{subfigure}

\caption{Comparison of convergence rate} \label{fig:1}
\end{figure}

Figure ~\ref{fig:1} compares the practical convergence rate. During the different ranges of iterations, solvers can have different performances. For example, normally FISTA is faster within the first ten iterations but only reaches the optimal solution in the experiment of dimensionality reduction. When loss function and regularizer are both smooth as in Kernel SVM, FISTA shows exactly convergence rate $O(1/n^2)$. If we prolong the line of FISTA in Figure ~\ref{fig kernel svm}, we can see FISTA needs about $10^{7.8}$ iterations to reach the optimal solution. However, PD CP only needs about $10^4$ iterations. For the last three experiments where loss functions and regularizers are both non-smooth, Fobos shows a convergence rate $O(1/\sqrt{n})$. But for all experiments, PD CP is the fastest to reach the optimal solution. Although the performances of PD CP with different values of $a$ are different, they show a similar practical convergence rate as shown in Figure ~\ref{fig cs} and ~\ref{fig linear svm}. PD CP has a much better practical convergence rate which is even better than $O(1/n^2)$. The experimental results show that FISTA is less capable to handle the case of two non-smooth terms. Furthermore, we may not get the optimal values by FISTA since we use the smoothing techniques. In all experiments, Online PD CP is better than or equal with PD CP. Refer the supplementary material for more results of machine learning problems. From Table ~\ref{tab rt}, we can observe PD CP is still very competitive. Only in Multi-Task Learning, the running time per iteration of PD CP is slower than OSGA's. However, by observing Figure ~\ref{fig MTL c}, PD CP needs much less number of iterations to approach the optimal solution. Overall, PD CP have the superior performance among all machine learning problems considered in this paper.

\section{Conclusion}

This paper compares the performance of different optimization algorithms applied to six benchmark problems of machine learning. The primal dual algorithm \cite{raey} has the best perform with a fast empirical convergence rate in all problems concerned in this paper. Moreover, we give two heuristic suggestions for the primal dual algorithm. We hope that machine learning problems can make good use of the progress in optimization. When we use an optimization algorithm to solve a machine learning problem, we need to set the value of parameters both in machine learn problem and optimization algorithm. Our future concern is how to set them automatically. Our experiments show that PD CP is an efficient and robust solver for machine learning problems. Future work is to give theoretical explanation about the empirical convergence rate of PD CP and the convergence of Online PD CP.

\begin{table}[H] 
%\vspace{0.3cm}
\noindent\begin{tabularx}{\textwidth}{cccc}
    \toprule
   per iteration (s)                      &\vtop{\hbox{\strut Dimensionality}\hbox{\strut Reduction}}   & Linear SVM  &Kernel SVM  \\\midrule
PD CP   & $4.164\times 10^{-4}$      & $11.1427\times 10^{-3}$  & $0.1009$ \\
OSGA   &  $6.25\times 10^{-4}$       &   $12.1379\times 10^{-3}$ &$0.3$   \\
FISTA   &  $5\times 10^{-4}$         &  $21.5387\times 10^{-3}$ & $ 0.1052$ \\\bottomrule
  per iteration (s)     & Feature Selection &\vtop{\hbox{\strut  Multi-Task}\hbox{\strut Learning}}   &Matrix Factorization   \\\midrule
  PD CP   & $1.7648\times 10^{-2}$   & $0.6214\times 10^{-3}$  & $4.69386$ \\
OSGA   &  $4.4\times 10^{-2}$    &   $0.3381\times 10^{-3}$ &$16.80799 $   \\
FISTA   &  $6.4\times 10^{-2}$       &  $0.8014\times 10^{-3}$ & $11.113636$ \\\bottomrule
\end{tabularx}\vspace{-0.3cm}\caption{Running time}\label{tab rt}
\end{table}

\iffalse
\section*{Acknowledgments}

This work was supported by whomever. We thank Prof.~X for her
thoughtful comments.
\fi

\bibliography{klkl}

\begin{thebibliography}{10}

\bibitem{Beck:2009:FIS:1658360.1658364}
Amir Beck and Marc Teboulle.
\newblock A fast iterative shrinkage-thresholding algorithm for linear inverse
  problems.
\newblock {\em SIAM J. Img. Sci.}, 2(1):183--202, March 2009.

\bibitem{Bennett:2006:IOM:1248547.1248593}
Kristin~P. Bennett and Emilio Parrado-Hern\'{a}ndez.
\newblock The interplay of optimization and machine learning research.
\newblock {\em J. Mach. Learn. Res.}, 7:1265--1281, December 2006.

\bibitem{raey}
Antonin Chambolle and Thomas Pock.
\newblock A first-order primal-dual algorithm for convex problems with
  applications to imaging.
\newblock {\em Journal of Mathematical Imaging and Vision}, 40(1):120--145,
  2011.

\bibitem{Chapelle2007}
Olivier Chapelle.
\newblock Training a support vector machine in the primal.
\newblock {\em Neural Computation}, 19:1155--1178, 2007.

\bibitem{Duchi:2009:EOB:1577069.1755882}
John Duchi and Yoram Singer.
\newblock Efficient online and batch learning using forward backward splitting.
\newblock {\em J. Mach. Learn. Res.}, 10:2899--2934, December 2009.

\bibitem{Gao:2012:DRV:2206432.2206550}
Junbin Gao, Qinfeng Shi, and Tib{\'e}rio~S. Caetano.
\newblock Dimensionality reduction via compressive sensing.
\newblock {\em Pattern Recogn. Lett.}, 33(9):1163--1170, July 2012.

\bibitem{Osga}
Arnold Neumaier.
\newblock Osga: A fast subgradient algorithm with optimal complexity.
\newblock Unpublished manuscript, 2014.

\bibitem{ICML2013_pele13}
Ofir Pele, Ben Taskar, Amir Globerson, and Michael Werman.
\newblock The pairwise piecewise-linear embedding for efficient non-linear
  classification.
\newblock In {\em ICML}, 2013.

\bibitem{DBLP:conf/icml/YangXKL10}
Haiqin Yang, Zenglin Xu, Irwin King, and Michael~R. Lyu.
\newblock Online learning for group lasso.
\newblock In {\em ICML}, 2010.

\bibitem{Yangarxivpd}
Tianbao Yang, Mehrdad Mahdavi, Rong Jin, and Shenghuo Zhu.
\newblock An efficient primal-dual prox method for non-smooth optimization.
\newblock {\em Mach Learn}, 2014.

\end{thebibliography}
\end{document}